\documentclass[review]{elsarticle}

\usepackage{amssymb}

\usepackage{makecell}
\usepackage{ragged2e}
\usepackage{pifont}
\usepackage{float}
\usepackage{color}
\usepackage{amsmath}
\usepackage{color}
\usepackage{algorithm}
\usepackage{amssymb}
\usepackage{algorithmicx}
\usepackage{algpseudocode}
\usepackage{array}
\usepackage{amsfonts}
\usepackage[caption=false,font=normalsize,labelfont=sf,textfont=sf]{subfig}
\usepackage{wrapfig} % 用于插入biography
\usepackage{flushend} % 用于让最后一页的biography以行对齐

\usepackage{fixltx2e}

\usepackage{stfloats}

\usepackage{url}
 \usepackage{bm}
\usepackage{caption}
\usepackage{multirow}
\usepackage{booktabs}
\usepackage{enumitem}
\usepackage{amsmath}
\usepackage{bbding}
\usepackage{array}
\newtheorem{definition}{Definition}

\journal{Pattern Recognition}

\begin{document}

\begin{frontmatter}

%% Title, authors and addresses

%% use the tnoteref command within \title for footnotes;
%% use the tnotetext command for theassociated footnote;
%% use the fnref command within \author or \address for footnotes;
%% use the fntext command for theassociated footnote;
%% use the corref command within \author for corresponding author footnotes;
%% use the cortext command for theassociated footnote;
%% use the ead command for the email address,
%% and the form \ead[url] for the home page:
%% \title{Title\tnoteref{label1}}
%% \tnotetext[label1]{}
%% \author{Name\corref{cor1}\fnref{label2}}
%% \ead{email address}
%% \ead[url]{home page}
%% \fntext[label2]{}
%% \cortext[cor1]{}
%% \affiliation{organization={},
%%             addressline={},
%%             city={},
%%             postcode={},
%%             state={},
%%             country={}}
%% \fntext[label3]{}

\title{Topology Reorganized Graph Contrastive Learning with Mitigating Semantic Drift}

%% use optional labels to link authors explicitly to addresses:
%% \author[label1,label2]{}
%% \affiliation[label1]{organization={},
%%             addressline={},
%%             city={},
%%             postcode={},
%%             state={},
%%             country={}}
%%
%% \affiliation[label2]{organization={},
%%             addressline={},
%%             city={},
%%             postcode={},
%%             state={},
%%             country={}}

\author[mymainaddress,mysecondaryaddress]{Jiaqiang Zhang }
\ead{zhangjq@nuaa.edu.cn}

\author[mymainaddress,mysecondaryaddress]{Songcan
	Chen\corref{mycorrespondingauthor}}
\cortext[mycorrespondingauthor]{Corresponding author}

\address[mymainaddress]{College of Computer Science \& Technology, Nanjing University of Aeronautics \& Astronautics,
	Nanjing, Jiangsu, 211106, China}
\address[mysecondaryaddress]{MIIT Key Laboratory of Pattern Analysis and Machine Intelligence, Nanjing University of Aeronautics \& Astronautics, Nanjing, Jiangsu, 211106, China}

\begin{abstract}
Graph contrastive learning (GCL) is an effective paradigm for node representation learning in graphs. The key components hidden behind GCL are data augmentation and positive-negative pair selection. Typical data augmentations in GCL, such as uniform deletion of edges, are generally blind and resort to local perturbation, which is prone to producing under-diversity views. Additionally, there is a risk of making the augmented data traverse to other classes. Moreover, most methods always treat all other samples as negatives. Such a negative pairing naturally results in sampling bias and likewise may make the learned representation suffer from semantic drift. 
Therefore, to increase the diversity of the contrastive view, we propose two simple and effective global topological augmentations to compensate current GCL. One is to mine the semantic correlation between nodes in the feature space. The other is to utilize the algebraic properties of the adjacency matrix to characterize the topology by eigen-decomposition. With the help of both, we can retain important edges to build a better view. 
To reduce the risk of semantic drift, a prototype-based negative pair selection is further designed which can filter false negative samples. 
Extensive experiments on various tasks demonstrate the advantages of the model compared to the state-of-the-art methods.

\end{abstract}

%%Graphical abstract
%\begin{graphicalabstract}
%\includegraphics{grabs}
%\end{graphicalabstract}

%%Research highlights
%\begin{highlights}
%\item Research highlight 1
%\item Research highlight 2
%\end{highlights}

\begin{keyword}
%% keywords here, in the form: keyword \sep keyword
Graph Neural Network, Self-Supervised Learning, Contrastive Learning, Node Representation Learning
%% PACS codes here, in the form: \PACS code \sep code

%% MSC codes here, in the form: \MSC code \sep code
%% or \MSC[2008] code \sep code (2000 is the default)

\end{keyword}

\end{frontmatter}

%% \linenumbers

%% main text
\section{Introduction}
\label{Introduction}

Graphs are capable of modeling complex interactions between objects that occur naturally in many real-world scenarios, such as in social networks, shopping websites, and citation networks \cite{9770382}. 
Learning and exploring the features of nodes and structures on these graphs facilitates various real-world challenges and applications. Graph Neural Networks (GNNs) and their various variants are a class of methods that can be used for representation learning on graph data. In recent years, they have achieved great success in dealing with graph analysis problems such as node classification and clustering \cite{welling2016semi,pan2023beyond,liu2023hard}, link prediction \cite{ZHANG2023109537}.

Most of these GNN methods adopt the paradigm of supervised learning, which extensively rely on label information to guide model learning \cite{zhang2019heterogeneous}. However, high-quality labeled data is laborious and expensive to collect in the real world. 
%Hence, self-supervised learning (SSL), a paradigm of exploring the knowledge inside unlabeled data in an unsupervised manner, is introduced for graph representation learning \cite{9770382}\cite{xie2022self}. 
%self-supervised learning (SSL) is a promising paradigm of exploring the knowledge inside unlabeled data in an unsupervised manner. 
Recently, self-supervised learning (SSL) is a promising paradigm for exploring the knowledge inside unlabeled data to alleviate the dependence on labeled data.
With the great success of SSL in computer vision \cite{chen2021exploring} and natural language processing \cite{gao2021simcse}, researchers also naturally apply SSL to graph-structured data \cite{li2023homogcl,yin2022autogcl}.
Existing SSL methods can be divided into three categories: predictive, generative, and contrastive \cite{xie2022self}. Predictive methods generate pseudo-labels with general prior knowledge, and design prediction-based auxiliary tasks to train \cite{hwang2020self}. Generative models generally regard the rich information in graph data such as structure and attribute information as self-supervised signals for reconstruction learning \cite{you2018graphrnn}. %Contrastive learning forms two views through augmentations, then pulls the defined positive pairs closer and the negative pairs farther away in the representation space \cite{qiu2020gcc}\cite{zhu2021graph}\cite{lee2022augmentation}, which is the focus of this paper.is one of the most successful
Contrastive learning is one of the most successful area in SSL \cite{lee2022augmentation}. It has achieved comparable or better results than supervised learning in representation learning task, which is the focus of this paper.

The main components in contrastive learning (CL) are augmentation, positive-negative pair selection, and contrastive objective \cite{zhu2021empirical}.
In particular, we first generate multiple contrastive views for each instance through data augmentations. Then the different views corresponding to the same instance are regarded as positive pairs, and all other instances usually negative samples. Finally, contrastive learning is achieved by optimizing the contrastive objectives which measure the agreement of positive and negative pairs \cite{xie2022self}. Inspired by previous CL methods, 
%graph contrastive learning (GCL) is also advancing rapidly.
Deep Graph Infomax (DGI) \cite{velickovic2019deep} as the first deep extension, applies the Infomax criterion and augmentation of shuffling node features to develop a GCL proposal.
GraphCL \cite{you2020graph} further takes augmentations like node dropping, edge perturbation, and subgraph sampling to generate the contrastive views.
%COSTA \cite{zhang2022costa} then further propose the hidden features augmentation schemes to preserve the intrinsic attributes of graphs. 
%Moreover, HomoGCL \cite{li2023homogcl} uses neighboring nodes with special significance for nodes to expand the positive set.

However, for contrastive learning, data augmentation and positive-negative pair selection are critical \cite{9770382}, yet still remain under-explored in the graph.
First, we argue that typical augmentations in GCL are generally blind (influence-agnostic) and tend to resort to local perturbation. 
Unlike the massive data augmentation techniques available in images and text, graph augmentation schemes are not so straightforward to define in CL methods because graphs are much more complex due to their non-Euclidean nature (i.e., topology). 
%Empirical studies \cite{zhu2021empirical} have also concluded that the performance of GCL is largely related to the choice of topology augmentation. 
Existing state-of-the-art approaches \cite{zhang2022costa,li2023homogcl,zhu2020deep} adopt the uniform addition and deletion of edges to augment the graph, but this blind approach risks deleting edges that will be influential for representation learning. GCA \cite{zhu2021graph} introduces certain prior information to remove edges, such as the degree centrality. However, this operation is local without considering the importance of the topology structure globally. Previous studies \cite{hassani2020contrastive} have shown that the higher-order neighboring information is beneficial for graph representation learning.
%The choice of augmented view controls the information captured by the learned representations, as the framework results in representations that focus on information between views \cite{xie2022self}. 
%Unlike the massive data augmentation techniques available in images and text, graph augmentation schemes are not straightforward to define in CL methods because graphs are much more complex due to their non-Euclidean nature (i.e., topology). Empirical studies \cite{zhu2021empirical} have also concluded that the performance of graph CL is largely related to the choice of topology augmentation.
%We argue that augmentations in graph CL are generally blind and resort to local perturbation.require sophisticated optimization
%For example, uniform addition and deletion of edges are adopted in \cite{you2020graph} and \cite{zhu2020deep}, but this blind approach risks deleting edges that will be important for downstream tasks. \cite{zhu2021graph} introduces certain prior information to adaptively remove edges. However, it will bring a certain inductive bias and is local without considering the importance of the topology structure globally. 
%The contrastive views formed by augmentations determine the information captured by the learned representations \cite{xie2022self}.
Some works \cite{you2021graph,kefato2021jointly} design interesting data-independent augmentation strategies, but they may incur sophisticated optimization, or fail to exploit task-related information in graph data due to the black-box nature.
If we want to thoroughly augment the topology so that the contrastive views can better reflect the characteristics of the graph, we deem there are two ways can go. One way is to find an indicator that globally quantifies the topological structure of a graph in order that those important parts can be augmented to form a new contrastive view.  In addition to the topological space, the structural information in the feature space is also important for the representation of the graph data \cite{wang2020gcn}.
Therefore, another way is not to adjust the original topological structure, but instead to take use of the feature structure of the data, such as using the feature similarities of the nodes to construct a global semantic relationship of the graph.
\begin{figure}[!t]
	\vspace{1.0em}
	\begin{center}
		\includegraphics[scale=0.6]{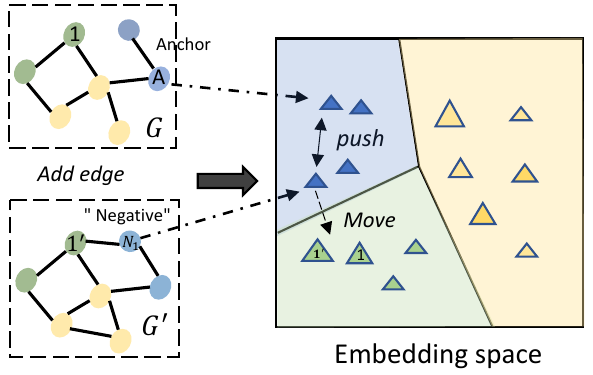}
		\caption{Illustration of the semantic drift problem} 
		\label{p1}
	\end{center}
	\vspace{-2.0em}
\end{figure} 

Additionally, in the positive-negative pair selection of GCL, there has the risk of making the representation of data traverse to other classes (ie. semantic drift). Most of the GCL models usually treat all other samples as negative without selection. 
%Negative examples obtained in this setting are not guaranteed to have truly different semantics \cite{robinson2020contrastive}. 
This setting may erroneously push some samples with similar semantics far away in the representation space, or even push them to other classes. Meanwhile, the risk of semantic drift will also occur in the graph augmentation stage, which will bring more serious consequences when combined with the uniform selection of negative samples. 
Take Figure (\ref{p1}) as an example, graph $\mathcal{G^{'}}$ is the augmented view of $\mathcal{G}$ by adding edges. 
Following the general contrastive learning paradigm, when node A is regarded as an anchor, although node $ N_{1} $ has the same label as node $ A $, it is also regarded as its negative sample being separated in the representation space by contrastive objective. Since node $ N_{1} $ and the samples of the green class have a connection relationship at this time, the obtained representation after GNN will gather some information about the green class. Therefore, node $ N_{1} $ is erroneously pushed away from its peers while pulling it toward the green class in the embedding space. 
%the final learned node representation will be less discriminative. 
Although some works have noticed the phenomenon of semantic shift, such as \cite{lin2022prototypical,peng2022graph}, they focus on the problem at the graph level and do not point out further problems caused by the combination of data augmentation.

To alleviate the above drawbacks, we propose a novel \underline{\textbf{Graph}} contrastive learning approach with \underline{\textbf{T}}opology reorganized augmentation and a \underline{\textbf{P}}rotoyp-based negative pair selection (\textbf{GraphTP} for convenience). 
First, to generate more informative contrastive views,  
we focus on the design of graph augmentations with the topological reorganization. 
In detail, we propose two ways. One is to focus on the adjacency matrix, quantify the topology globally and then augment it. 
That is, the matrix is converted into eigenvalues and eigenvectors by eigen-decomposition. 
In simple terms, the eigenvectors can be regarded as the features corresponding to the matrix, and the corresponding eigenvalues indicate how important the features are. 
%An intuitive idea is that important parts need to be reinforced, i.e. the weights of leading features corresponding to relatively larger eigenvalues should be inflated. 
So we exponentiate the eigenvalues to highlight important connections of the graph and then calculate with the eigenvectors to restore a new adjacency matrix.
The other is to use the global semantic structure information in the feature space. 
We construct a structure by calculating the feature similarity between nodes and then finding the $ k $ most similar nodes in for each node as its important neighbors. 
Second, for the risk of semantic drift in learned representations, 
we attempt to select high-quality negative samples that are semantically correct. Specifically, for each node, the semantic similarity of its corresponding prototype with the candidate negative sample is calculated by cosine. If the semantic similarity of the pair is high (low), the Bernoulli sampling will be conducted to discard (keep) it. 

In this paper, the main contributions are summarized as follows:

$ \bullet $ We thoroughly augment the graph structure from two perspectives, i.e., finding the important parts in graph globally in the topology and the feature space, which aims to form a more informative contrastive view to reflect the depth-related information inside the graph data.

$ \bullet $  We design a prototype-based negative pair selection strategy to reduce the risk of semantic drift, which can filter out more precise negative pairs with really different semantics. 
%Meanwhile, we theoretically analyze the advantages of current pair selection methods from the view of gradient descent.

$ \bullet $  We conduct extensive experiments on five real-world datasets compared to state-of-the-art methods. The results verify the effectiveness of our proposal on the tasks of node classification, clustering, and similarity search.

\section{RELATED WORK}
In this section, we review the related work from three aspects: graph neural networks, unsupervised graph representation learning, and graph contrastive Learning.
\subsection{Graph Neural Networks}
Graph exists widely in the real world. Graph neural networks (GNNs), which learn graph embeddings by using attribute features and topological information, have been extensively studied \cite{bruna2013spectral}. GNNs usually adopt the message passing paradigm, that is, iteratively updates the representation of nodes by aggregating the representations of their neighbors, and sums up the representation of nodes through pooling operations to obtain the representation of the entire graph. 
GNN is first introduced in \cite{bruna2013spectral}, which combined graph Laplacian to design a graph convolution operation in the Fourier domain. Then, GCN \cite{welling2016semi} builds a bridge between the spectral domain and the spatial domain in GNN. It uses the first-order Chebyshev polynomial filter approximation for efficiency and only aggregates node features from first-order neighbors each time. GAT \cite{velivckovic2017graph} further introduces the attention mechanism to consider the importance of different node neighbors instead of simple aggregation. GraphSAGE \cite{hamilton2017inductive} provides four functions for aggregating nodes: mean/max/LSTM/Pooling. KerGNNs \cite{feng2022kergnns} combines the graph kernel method, which naturally extends the CNN framework to the graph, and brings a certain degree of interpretability.
Meanwhile, existing GNN methods have been successfully applied in various fields with great success, such as anomaly detection \cite{KIOUCHE2021107746} and node clustering \cite{liu2023hard,pan2023beyond}. 
But most of GNNs are to learn the representation of nodes in an end-to-end manner under the paradigm of supervised learning. %relying heavily on the guidance of label.
\subsection{Unsupervised Graph Representation Learning}
Unsupervised graph representation learning aims to learn node embeddings on the graph when data labels are not available. Early methods focus more on using the structural information of the graph to learn the representation, such as random walk based and kernel-based methods \cite{perozzi2014deepwalk,grover2016node2vec,yanardag2015deep}. 
The former method takes walks across nodes randomly on the graph and then flattens the graph into a sequence for learning.
Node2vec \cite{grover2016node2vec} is a representative work among them, which can effectively explore different neighborhoods by designing a biased random walk function. Kernel-based methods, such as Graphlet kernels \cite{yanardag2015deep}, use the dependency information between substructures and then combined it with the kernel function to give the similarity between graphs for representation learning. But they all cannot simultaneously make use of node features and topology information \cite{hassani2020contrastive}. Recently, the representation learning of data through well-designed pretext tasks without labels has received extensive attention, which can be roughly divided into three categories: contrastive, predictive, and generative \cite{zhu2020deep}. The contrastive-based graph representation learning method will be expanded in the following part.
\subsection{Graph Contrastive Learning}
Recently, graph contrastive learning has been extensively studied due to the relatively large success of self-supervised contrastive learning in computer vision and natural language processing \cite{guo2022hcsc}. General graph contrastive learning can be roughly divided into three modules: contrastive objective, data augmentation, and positive-negative sample pair selection. The existing work is mostly based on the innovation of three modules \cite{9770382,LIU2024109907}.
Deep Graph InfoMax (DGI) \cite{velickovic2019deep} firstly applies the Infomax criterion to the graph, and proposes to compare the node representation derived from the corrupted graph with the whole graph representation.
For the design of data augmentation, MVGRL \cite{hassani2020contrastive} learns node-level and graph-level representations by injecting global structural information into the graph to obtain a contrastive view. GraphCL \cite{you2020graph} and GRACE \cite{zhu2020deep} use the SimCLR \cite{chen2020simple} framework to propose a variety of heuristic graph data augmentation methods, such as masking node features, discarding nodes, and removing edges .etc. GCA \cite{zhu2021graph} makes a further improvement by introducing prior information, such as node-based degrees, to adaptively augment the graph. 
COSTA \cite{zhang2022costa} focuses on hidden feature augmentation to avoid getting a biased node representation. 
Meanwhile, AutoGCL\cite{yin2022autogcl} has turned to the use of model augmentation to mitigate the risk of semantic destruction during data augmentation, but the semantic drift in positive-negative pair selection still exists. ProGCL \cite{xia2022progcl} mines the hard negative samples by the technology of mixup.
Besides, HomoGCL\cite{li2023homogcl} is the recent work that exploit the homophily assumption to complement graph contrastive learning.

In this paper, we consider the guided augmentation for topology which is information specific to the graph, and then improve the instance-based contrastive learning by designing a new negative sample selection strategy to alleviate the problem of semantic drift.
\section{THE PROPOSED FORMULATION}
In this section, the terminology and problem definitions are given as follows. 
Let $\mathcal{G}=(\mathbf{X}, \mathbf{A})$ denote an undirected graph, where $ \textbf{A} $$ \in \mathbb{R}^{N \times N} $ is the adjacency matrix, $N$ is the number of the nodes.                   
$ \mathbf{A}_{i,j} =1$ if there is an edge between node $ v_{i} \in \mathcal{V} $ and $ v_{j} \in \mathcal{V}  $ and $ \textbf{A}_{i,j} =0$ otherwise, where $\mathcal{V}$ is the set of $ N $ nodes.
$ \mathbf{X} =[\mathbf{x}_{1}, \mathbf{x}_{2},..., \mathbf{x}_{N}]^{T}$ $ \in $ $ \mathbb{R}^{N \times d} $ is the feature matrix, where $ \mathbf{x}_{i} $ is the $ \textit{i} $-th row of $\mathbf{X} $  and denotes the feature vector of $ v_{i} $.  The unnormalized graph Laplacian of $\mathcal{G}$ is $\mathbf{L}=\textbf{D}-\textbf{A}$, where $  \textbf{D}=diag(d_{1}, d_{2},..., d_{n}) $ is the degree matrix of $\textbf{A}$ and  $ d_{i} = \sum_{j \in \mathcal{V} }\textbf{A}_{i,j}$. When the adjacency matrix $\textbf{A}$ is normalized to ${\hat{\textbf{{A}}}}=\textbf{D}^{-\frac{1}{2}}\textbf{A}\textbf{D}^{-\frac{1}{2}} $, the normalized Laplacian matrix can also be defined as $\hat{\mathbf{L}}=\textbf{D}^{-\frac{1}{2}}\textbf{L}\textbf{D}^{-\frac{1}{2}}$.
In this paper, we take a GNN encoder \cite{welling2016semi} as the backbone network. 
\begin{definition}[Graph Neural Network]
	Given an graph $\mathcal{G}=(\mathbf{X}, \textbf{A})$, a typical graph neural network mainly consists of two components: Message Aggregation and Combine:
	\begin{align}
		\mathbf{m}^{(l)}_{v} &= AGGREGATE^{(l)}(	\mathbf{h}^{(l-1)}_{v^{'}}: v^{'} \in (\mathcal{N}(v )\cup v)) \nonumber, \\
		\mathbf{h}^{(l)}_{v} &= COMBINE^{(l)}(\mathbf{m}^{(l)}_{v}, \mathbf{h}^{(l-1)}_{v} ) \label{eq: gnn},
	\end{align}
	where $AGGREGATE$ and $COMBINE$ are message aggregation and combine functions. 	$ \mathbf{h}^{(l)}_{v} $ represents the embedding of node $v$ in the $l$-th layer, $ \mathcal{N}(v)  $ denotes the neighbor nodes of $v$. For each node, the $l$-hop neighbor information can be captured by stacking an $l$-layer GNN.
	%In general, a graph neural network generates the representation of a node by combining representation of its own and its neighbours.
\end{definition}

\begin{definition}[Instance-based contrastive learning]
	%A widely used self-supervised learning contrastive method \cite{zhu2021empirical} is to compare a positive pair with negative instance pairs. 
	Instance-based contrastive learning is one of the widely used paradigms for self-supervised learning (SSL).
	In principle, for a anchor node $ v_{i} $, the same node corresponding to other views is regarded as positive to be pulled closer, while all other instances as negative to be pushed away.  
	Specifically, given the representation of a positive pair ($ \mathbf{z}_{i},\mathbf{z}_{i}^{'} $), the agreement is maximized for this positive pair and minimized for negative pairs via the standard InfoNCE loss:
	\begin{equation}
		{l}(\mathbf{z}_{i}, \mathbf{z}_{i}^{'})=log\frac{e^{\theta(\mathbf{z}_{i},\mathbf{z}_{i}^{'})/\tau}}{\sum_{\mathbf{z}_{j} \in {\mathbf\{{z}_{i}^{'}} \cup \mathcal{B}(v_{i})\}}e^{\theta(\mathbf{z}_{i},\mathbf{z}_{j})/\tau}}
	\end{equation}
	\begin{equation}
		\mathcal{L}_{info}=-\frac{1}{2N}\sum^{N}_{i=1}({l}(\mathbf{z}_{i}, \mathbf{z}_{i}^{'})+l(\mathbf{z}_{i}^{'},\mathbf{z}_{i}))
		\label{l1}
	\end{equation}
	where $ N $ is the number of nodes, $ \mathcal{B}(v_{i}) $ is a set of negative instances for node $ v_{i} $, $ \tau $ is a
	temperature parameter.
\end{definition}

In this work, we focus on the graph representation learning without label supervision. That is, given an unlabeled graph $\mathcal{G}=(\mathbf{X}, \mathbf{A} )$ , where $ \mathbf{X} \in \mathbb{R}^{N \times d} $, 
$ \mathbf{A} \in \mathbb{R}^{N \times N}$, we aim to train a GNN-based encoder $ f(\cdot) $ to obtain high-quality low-dimensional node representations: $ f(\mathbf{X},\mathbf{A}) \in \mathbb{R} ^{N \times d^{'}} $ , ie. $ d^{'}\ll d $. Then these representations can be adopted in downstream tasks, such as node clustering, classification, and similarity search. 

\begin{table}[t]
	\tiny
	\centering
	\caption{Summary of the primary notations.} 
	\begin{tabular}{ p{62 pt}<{\centering} | p{159 pt}}  
		\toprule[1.0pt]
		Symbols & Description  \\
		\cmidrule{1-2}
		$\mathcal{G}=(\mathbf{X}, \mathbf{A})$ & A graph with feature matrix \\
		$\mathbf{X} \in \mathbb{R}^{N \times d}$ & The features matrix of $\mathcal{G}$ \\
		$\mathbf{A} \in \mathbb{R}^{N \times N}$ & The adjacency matrix of $\mathcal{G}$ \\
		$N$ & Number of nodes in $\mathcal{G}$\vspace{0.5mm}\\
		$\tau$ & The temperature parameter in GCL\vspace{0.5mm}\\
		$d$ & Number of dimensions of $\mathbf{X}$ \vspace{0.5mm}\\
		$d'$ & Number of dimensions of the latent representation \vspace{0.5mm}\\
		\cmidrule{1-2}
		${\mathcal{G}_{T}} = (\mathbf{X}, {\textbf{A}_{T}})$ &  The augmented version of the input graph $\mathcal{G}$ by topology reorganization \vspace{0.5mm}\\
		${\mathcal{G}^{'}} = ({\mathbf{X}_{1}}, {\textbf{A}^{'}})$ &  A contrastive view derived from $\mathcal{G}$ \vspace{0.5mm}\\
		${\mathcal{G}^{'}_{T}} = ({\mathbf{X}_{2}}, {\textbf{A}^{'}_{T}})$ &  A contrastive view derived from $\mathcal{G}_{T}$ \vspace{0.5mm}\\
		$k$ &  Top-$ k  $ strongly related neighbours for node $v$  \vspace{0.5mm}\\
		$K$ &  Number of cluster in the prototype-based negative sampler \vspace{0.5mm}\\
		\cmidrule{1-2}
		${\mathcal{G}^{'}} , {\mathcal{G}^{'}_{T}}$ &  The contrastive views of $\mathcal{G}$ \vspace{0.5mm}\\
		$f(\cdot)$ & The GNN encoder used in our model\\
		${\textbf{Z}_{1}} \in \mathbb{R}^{N \times d'}$ & Final embedding matrix for $\mathcal{G}^{'}$ \vspace{0.5mm}\\ 
		${\textbf{Z}_{2}} \in \mathbb{R}^{N \times d'} $ & Final embedding matrix for $\mathcal{G}^{'}_{T}$ \vspace{0.5mm}\\ 
		$\textbf{z}_{i}, \textbf{z}^{'}_{i} \in \mathbb{R}^{1 \times d'}$ &  The representation of node $v_{i}$ in $Z_1, Z_2$\vspace{0.5mm}\\
		$\mathbf{S}\in \mathbb{R}^{N \times N}$ &  The similarity matrix in feature space \vspace{0.5mm}\\
		$\textbf{B} \in \mathbb{R}^{N \times N}$ &  The new weighted adjacency matrix \vspace{0.5mm}\\
		$\mathcal{B}_{f}(v) $ &  The new negative sample set of node $ v $ \vspace{0.5mm}\\
		$\textbf{W}^l \in \mathbb{R}^{d \times d'}$ & The $l$-th layer trainable weight matrix in $f({\cdot})$\\
		\bottomrule[1.0pt]
	\end{tabular}
	\label{table:notation}
\end{table}

\section{METHODOLOGY}
\subsection{Overview}
\begin{figure*}
	\begin{center}
		\includegraphics[scale=0.5]{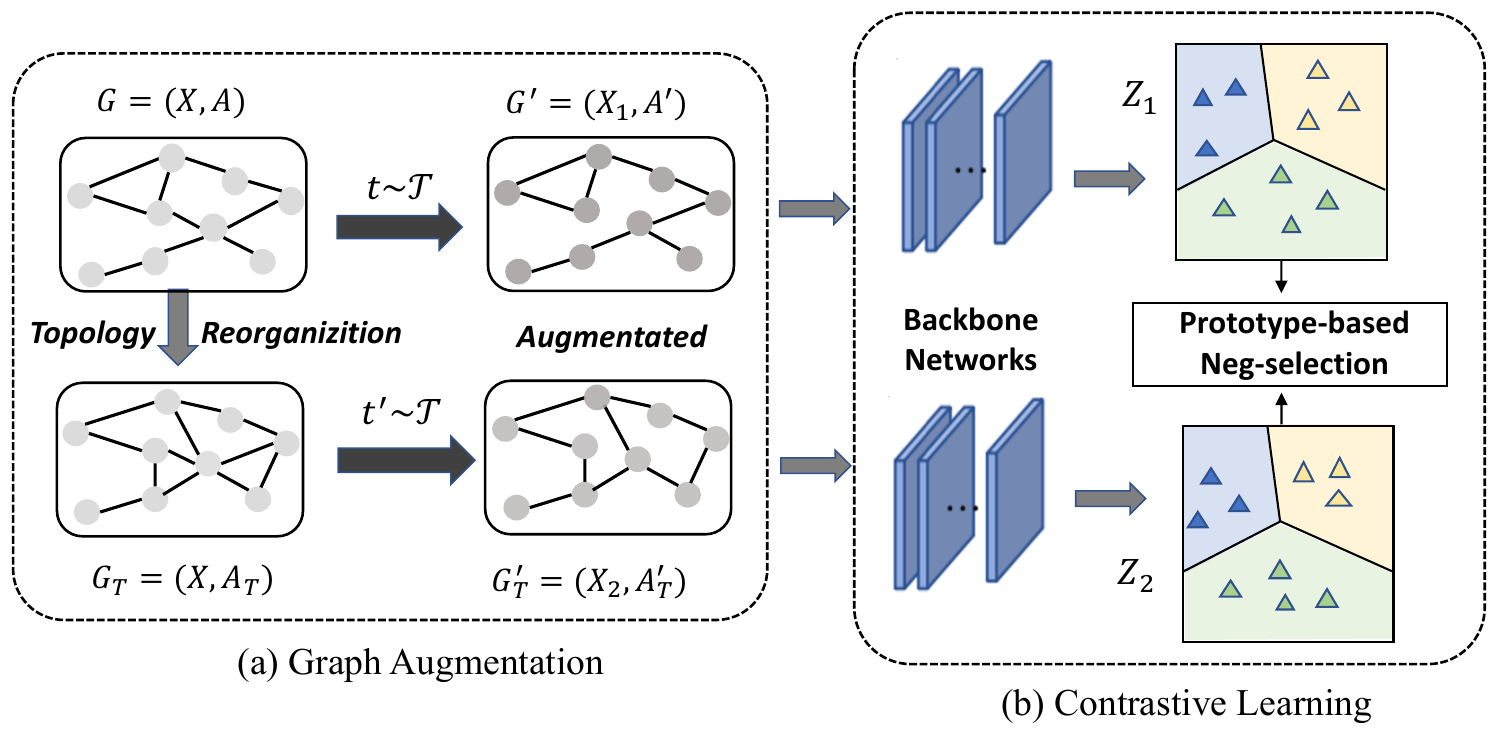}
		\caption{The framework of the proposed model. }
		%Given a graph $ \mathcal{G}(\mathbf{X},\mathbf{A}) $, we can first obtain its topology augmented graph $ \mathcal{G}_{T}(\mathbf{X},\mathbf{A}_{T}) $ by Sec.\ref{subsec:topology}. Furthermore, we define the set of typical  perturbations as $ \mathcal{T} $. 
		%Within every epoch, we sample two perturbations $ t \sim \mathcal{T}$ and  $ t^{'} \sim \mathcal{T} $ to generate two views: $ \mathcal{G}^{'}=t(\mathcal{G}) $  and $ \mathcal{G}^{'}_{T}=t^{'}(\mathcal{G} _{T})$. 
		%Then the two views are input into the encoder $ f $ to get the node embeddings: $ \mathbf{Z}_{1} $ and  $ \mathbf{Z}_{2} $. At this time, prototype-based negative pair selection is applied, for each node $ v_{i} $ in the view $ \mathcal{G}^{'} $, its corresponding node $ v_{i}^{'}  $  in another view $ \mathcal{G}^{'}_{T} $ is regarded as the positive. While its set of negative instances can be obtained by Sec. \ref{subsec:GCL}. }
		\label{model}
	\end{center}	
\end{figure*}

In this section, we introduce the proposed model GraphTP in detail. As shown in Figure.\ref{model}, our model is mainly divided into two steps, \textit{data augmentation with topology reorganization} and \textit{contrastive learning with prototype-based negative pair selection}. Specifically, we first augment the topology globally in a targeted manner to obtain an informative augmented view, for which we provide two schemes. (the left part in Figure. \ref{model}). In addition, feature masking and edge dropping are performed to further increase the diversity between views. Then, the two contrastive views $ \mathcal{G}^{’}(\mathbf{X}_{1}, \mathbf{A}^{'}) $ and $ \mathcal{G}_{T}^{’} (\mathbf{X}_{2}, \mathbf{A}^{'}_{T}) $ are input into the GNN-based encoder to obtain the representations: $ Z_{1} \in \mathbb{R}^{N \times d^{'}}  $, $ Z_{2} \in \mathbb{R}^{N \times d^{'}} $, respectively. At last, to alleviate the risk of semantic drift \cite{zhang2022costa} in the process of contrastive learning, for each node, we design a prototype-based negative sample selection strategy to select more accurate negative samples that have true semantics irrelevant to the query. The main steps of the model will be introduced: Sec.~\ref{subsec:topology} for the designed augmentation and Sec.~\ref{subsec:GCL} for the prototype-based sample selection.
\subsection{Graph Augmentation with Topology Reorganization}
In contrastive learning, the generation of "view" is a factor that controls the information captured by the representation \cite{9770382}. 
%The generation of "view" is an important factor in controlling the representational information \cite{oord2018representation}. 
So we should carefully design the augmentation in order that the generated views can reflect the depth-related information inside the data.
%To this end, we take use of the adjacency matrix and the semantic structure of feature space to .
%Since topology is crucial in graph representation learning \cite{gao2021self}, this motivates us to augment graphs from the perspective of topology. 
To this end, we design two schemes, one is the feature-space based, which makes use of the semantic structure in feature space, and the other is the matrix-transformation based, which utilizes the algebraic properties of the adjacency matrix.

%which are the construction of the semantic relationship based on the node feature space and the matrix transformation based on the eigenvalues of the adjacency matrix. Next, we will introduce them in detail, respectively.
\label{subsec:topology}
\subsubsection{Scheme 1: Feature-Space Based }
\label{subsec:s1}

In real-world scenarios, the correlation between graph and downstream tasks is usually very complex and can be related to its topology or node features, or their combination. Therefore, the implicit relationship between the node in feature space is also important and can be exploited \cite{wang2020gcn}. 

In detail, we build a graph $ \mathcal{G}_{f}(\mathbf{A}_{f},\mathbf{X}) $ based on node features. 
Given the feature matrix of nodes: $ \mathbf{X} $, we first calculate the feature similarity between nodes to generate a similarity matrix $ \mathbf{S}$. The similarity here can be measured by various methods. We list three commonly used methods: 
1) Heat Kernel, 2) Mahalanobis Distance, 3) Cosine Similarity. 
Here the Cosine Similarity is chosen because of its good boundedness and measure invariance.
\begin{equation}
	\mathbf{S}_{ij}=\frac{\mathbf{x}_{i} \cdot \mathbf{x}_{j}}{|\mathbf{x}_{i}||\mathbf{x}_{j} |}.
\end{equation}
where $ \mathbf{x}_{i} $ and $\mathbf{x}_{j} $ is the feature vectors corresponding to  $ v_{i} $ and $ v_{j} $.
According to the $ \mathbf{S}$, we select the top-$ k $ nodes with high similarities as neighbors for each node to finally construct the new adjacency matrix $ \mathbf{A}_{f} $.

\subsubsection{Scheme 2: Matrix-Transformation Based}
\label{subsec:s2}
For the augmentation of topology structure, different from the previous random and local addition and deletion of edges, we aim to quantify the entire topological relationship by its algebraic properties and then augment.
With the help of matrix and spectral theory \cite{welling2016semi}, the specific operation is as follows:

Given the graph $\mathcal{G}=(\mathbf{A}, \mathbf{X} )$, we first obtain the unnormalized graph Laplacian ${\mathbf{L}}=\mathbf{D}-\mathbf{A}  $, where the $ \mathbf{D} $ is the degree matrix of $ \mathbf{A} $ and $ d_{i}=\sum_{j \in \mathcal{V} } \mathbf{A}_{ij} $. So the symmetric normalized graph Laplacian can be denoted to $ \hat{\mathbf{L}}=\mathbf{D}^{-\frac{1}{2}}\mathbf{L}\mathbf{D}^{-\frac{1}{2}}$, its eigen-decomposition can be formulated as:
%$ \hat{\mathbf{L}}=\mathbf{I}_{n}-\hat{\mathbf{A}} $
\begin{equation}
	\hat{\mathbf{L}}=\mathbf{U}\Lambda\mathbf{U}^{T},
	\label{laa1}
\end{equation}
where $  \Lambda=diag(\lambda_{1},...,\lambda_{N}) $ and $  \mathbf{U}=[ \mathbf{u_{1}},..., \mathbf{u_{N}}] $ are the eigenvalues and corresponding eigenvectors of $ \hat{\mathbf{L}} $, respectively \cite{welling2016semi} . Further, assuming the rank of A to be $ N $, we have:
\begin{equation}
	\hat{\mathbf{L}}=\mathbf{U}\Lambda\mathbf{U}^{T}=\sum_{i=1}^N \lambda_{i}\mathbf{u}_{i}\mathbf{u}_{i}^{T},
	\label{laa}
\end{equation}
where $ \mathbf{u}_{i}\mathbf{u}_{i}^{T}, i=1,2,..., $  form a set of orthonormal base matrices for the topology space $ \mathbf{R}^{N \times N} $.

Generally speaking, the matrix $  \mathbf{u}_{i}\mathbf{u}_{i}^{T} $ corresponding to the larger eigenvalue $ \lambda_i $ always
indicates the the principal component of the topology space \cite{klicpera2019diffusion}. 
%The weights of the leading base graph $ \mathbf{u_{i}}\mathbf{u_{i}}^{T} $ corresponding to large eigenvalues $ \lambda_{i} $  to be inflated by exponentiation.
%We aim to obtain a new weight adjacency matrix, and its corresponding value is the degree of association between each node. 
%Thus, we process the eigenvalues of the Laplacian matrix.
Menwhile, inspired by the previous work \cite{liu2008fractional}, if we augment the larger eigenvalue with the exponentiation, the topological relationship in the graph can be augmented globally. 
For each node, the weights of influential edges that control the structural properties \cite{chung1997spectral} of graph are increased, and vice versa. 
Specifically, for the Laplacian matrix $ \hat{\mathbf{L}}$ in Eq. \ref{laa1}, the augmented matrix is defined as:
\begin{equation}
	{\mathbf{B}}=\mathbf{U}\Lambda^{\alpha}\mathbf{U}^{T},
\end{equation}
where $ \alpha $ is a tunable hyperparameter. Similar to the way in Sec.~\ref{subsec:s1}, with the help of $ {\mathbf{B}} $, for each node, we find its neighbors  with the top-$ k $ edge weights to construct a perturbed adjacency matrix $ \mathbf{A}_{p} $. At the same time, this operation can also be regarded as the augmentation of the low- and high-frequency information in the graph. Since $  0\leq \lambda_{1}\leq...\leq\lambda_{N}<2 $, and the larger eigenvalue always corresponds to high-frequency information \cite{chung1997spectral}, performing a power operation of $ \alpha>1 $ is equivalent to highlighting high-frequency information and suppressing low-frequency information to a certain extent, and the opposite is true when $ \alpha<1. $

%Meanwhile, we can find this operation is also equivalent to doing exponentiation on the Laplacian matrix:
%\begin{equation}
%	\hat{\mathbf{L}}^{\alpha}=\underbrace{(\mathbf{U}\Lambda\mathbf{U}^{T}) \mathbf{U}\Lambda\mathbf{U}^{T}...(\mathbf{U}\Lambda\mathbf{U}^{T})}_{\alpha}=\mathbf{U}\Lambda^{\alpha}\mathbf{U}^{T}={\mathbf{B}}
%\end{equation}
%where $ \mathbf{U}^{T} =\mathbf{U}^{-1}$, $ \mathbf{U}^{T}\mathbf{U}=\mathbf{U}^{-1}\mathbf{U}=\mathbf{I} $, $\mathbf{I} $ is the identity matrix.
%From the perspective of topology relationship propagation, the $ k $-th power operation on the matrix is equivalent to $ k $-time weight propagation of nodes along edges \cite{ding2021inductive}. Thus, we can also consider the benefits of this scheme from the perspective of propagation. The weighted matrix obtained by this scheme can represent the global correlation between nodes to a certain extent, for which the edges connected by influential nodes will have greater weight. 

Through the selection of these two schemes, we have two contrastive views : the graph $ \mathbf{G}(\mathbf{X},\mathbf{A}) $ and the topology augmented graph $ \mathbf{G}_{T}(\mathbf{X},\mathbf{A}_{T}) $, where the new adjacency matrix $ \mathbf{A}_{T} $ is chosen from $ \mathbf{A}_{f} $ or  $  \mathbf{A}_{p} $.  In order to further increase the diversity between views, we introduce the following augmentations \cite{zhu2021graph}: feature masking and the removal of edges, which are denoted to  $ \mathcal{T} $. 
The detailed operations are introduced as follows:

\vspace{0.6em}
1) \textbf{Feature Masking}:
At the node feature level, we add noise by using the random masking to some fractions of dimensions in node features, which is similar to salt-and-pepper noise in digital image processing. 

Specifically, we first define a masking vector $ \mathbf{m} \in \{0,1\}^{d} $ with the same dimension as the node feature vector, in which each element is independently sampled from a Bernoulli distribution with probability $ p_{i} $, ie., $ \mathbf{m}_{i} \sim$$Bernoulli(1-p_{i})$. The probability $ p_{i}  $ is calculatied by the method proposed by \cite{zhu2021graph}. The central idea of this calculation is that the probability $ p_{i}  $ reflects the importance of the $ i $-th dimension of node features.  

Taking the view-1 as an example, the original feature matrix $ \mathbf{X} $ with the augmentation of masking can be formulated as :
\begin{equation}
	{\mathbf{X}_{1}}=[\mathbf{x}_{1} \odot \mathbf{m},\mathbf{x}_{2} \odot \mathbf{m},...,\mathbf{x}_{n} \odot \mathbf{m} ],
\end{equation}
where the $  \odot  $ is the element-wise multiplication, $ \mathbf{X}_{1} $ is the augmented feature matrix.

\vspace{0.6em}
2) \textbf{Edge Removal}: For the edge augmentation, we adopt a similar scheme to the feature masking. For the edge $(i,j) $ between nodes $ i $ and $  j$, its discarded probability value $ p_{uv} $ is calculated by \cite{zhu2021graph}, and then is set as the parameter of the Bernoulli distribution for sampling. Formally, an edge masking vector is defined as $ \mathbf{m}^{e}$. Then the element $ \mathbf{m}_{uv} $ in $ \mathbf{m}^{e} $ can be calculated as follows: 
\begin{equation}\mathbf{m}_{uv}=
	\begin{cases} 
		Bernoulli(1-p_{uv})&  {\mathbf{A}_{ij}=1}\\
		0& {\mathbf{A}_{ij}=0}
	\end{cases}
\end{equation}

After the above operations, we have the view-$ 1 $:   $ \mathbf{G}^{'}(\mathbf{X}_{1},\mathbf{A}^{'}) $  and the view-$ 2$: $ \mathbf{G}^{'}_{T}(\mathbf{X}_{2},\mathbf{A}^{'}_{T}) $.
\begin{figure}[!t]
	\vspace{1.0em}
	\begin{center}
		\includegraphics[scale=0.6]{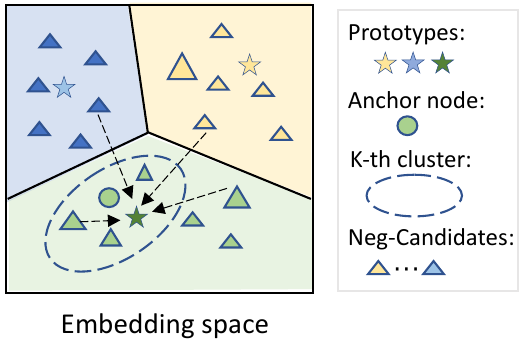}
		\caption{Illustration of the prototype-based negative pair selection} 
		\label{pp}
	\end{center}
%	\vspace{-2.0em}
\end{figure} 
\subsection{Prototype-based negative pair selection}
\label{subsec:GCL}
The general graph contrastive learning methods adopt the instance-based learning paradigm. Specifically, for an anchor node, all other instances are uniformly regarded as negative and pulled apart in the representation space  \cite{zhu2021empirical}. 
However, this simple approach cannot guarantee that these negative instances have different semantics from the anchor, resulting in a risk of semantic drift in node representations, as illustrated in Figure. \ref{p1}. 
To alleviate this risk, we propose a new strategy for negative pair selection to choose the instances whose semantics are irrelevant with the anchor, and show it in Figure. \ref{pp}. Inspired by \cite{liprototypical}, this semantics is measured by the feature similarity of each instance with the prototype of the anchor.
%We aims to select more precise negative instances with truly unrelated semantics. 
%That is, for any anchor node $ v $, we select more precise negative instances by performing Bernoulli sampling on its negative candidate set $\mathcal{B}(v) $. 
%Under such a sampling process, we would like to eliminate the instances in the $\mathcal{B}(v) $ that have highly similar semantics to the anchor, while retaining instances with low semantic relevance. Following the \cite{liprototypical}, the semantics are measured by calculating the feature similarity of each instances with the prototype of the anchor.
%For any anchor node $ v $ and the candidate set  $\mathcal{B}(v) $ of its negative instances, we calculate the feature similarity between each instance in the $\mathcal{B}(v) $ and its prototype.
%Then the similarity as the parameter of a Bernoulli sampling to determine whether to discard from $\mathcal{B}(v) $.
Specifically, for a given anchor $v$ and its corresponding representation $ \mathbf{z}  \in \mathbb{R}^{1 \times d'} $ in graph $ \mathcal{G} $, we first define the semantic similarity between it and prototype $ c $ :
\begin{equation}
	s( \mathbf{z} , c)=\frac{\mathbf{z}\cdot{c}}{\xi_{c}}
\end{equation} 
where the $ \cdot $ denotes the dot product, $ c \in C=\{c_{i}\}_{i=1}^{K} $, $C$ is generated by $ K $-means on the all node representations.
%To avoid the collapse of most nodes into a single cluster,
$ {\xi_{c}}  $ is the concentration of sample distribution around the prototype, and formulated as:
\begin{equation}
	\xi_{c}=\frac{\sum_{\mathbf{z}_{i} \in \mathbf{Z}_{c}\\}||\mathbf{z}_{i}-c||_{2} }{|\mathbf{Z}_{c}|log(|\mathbf{Z}_{c}|+\epsilon)}
\end{equation}
where $ \epsilon $ is a smooth parameter that balances the concentration between different clusters, avoiding small clusters with too large concentration \cite{liprototypical}. $ \mathbf{Z}_{c} $ is the set of node representations within cluster $ c $, where $ c \in \{c_{i}\}_{i=1}^{K}  $.

Thus, the prototype corresponding to this anchor $ v $ can be denoted as:
\begin{equation}
	c( \mathbf{z} )=argmax_{c \in \{c_{i}\}_{i=1}^{K}} s( \mathbf{z} ,c)
\end{equation}
At last, we could conduct the negative instance selection. For a candidate $  \mathbf{z}_{j}  \in \mathbb{R}^{1 \times d'}$ in the negative candidate set $ B(v) $ of anchor $ v $, we choose it if its semantic similarity to $ c( \mathbf{z} )  \in \mathbb{R}^{1 \times d'} $ is smaller than than those of other prototypes,
so its selected probability can be defined as:
\begin{equation}
	p( \mathbf{z} , \mathbf{z} _{j})=1-\frac{exp[s(\mathbf{z}_{j},c( \mathbf{z} ))]}{\sum_{i=1}^{K}exp[s(\mathbf{z}_{j},c_{i})]}.
\end{equation}
On such a basis, a Bernoulli sampling with the probability is performed on each negative candidate to obtain a more accurate set $ B_{f}(v) $:
\begin{equation}
	\mathcal{B}_{f}(v)=\{Bernoulli(p(\mathbf{z},\mathbf{z}_{j}))| \mathbf{z}_{j} \in \mathcal{B}(v)\}.
	\label{b}
\end{equation}

\renewcommand{\algorithmicrequire}{ \textbf{Input:}}     %Use Input in the format of Algorithm
\renewcommand{\algorithmicensure}{ \textbf{Output:}}    %UseOutput in the format of Algorithm
\begin{algorithm}[t!]
	\caption{The Overall Procedure of GraphTP} % 名称
	\begin{algorithmic}[1]
		\Require
		The graph: $\mathcal{G}( \mathbf{X},\mathbf{A})$ ;
		Topology augmented graph: $\mathcal{G}_{T}( \mathbf{X},\mathbf{A}_{T})$ (generated by Sec.\ref{subsec:topology});
		The temperature: $\tau$;
		Maximum training epochs: $ E$;
		Warm-up epoch: $T^{'}$;
		GNN-based encoder: $f$;
		The set of typical perturbations: $ \mathcal{T} $
		\Ensure
		Trained GNN $f$  ;
		\For{$epoch = 1$; $i<E$; $i++$ }
		\State Draw two perturbation functions $ t \sim \mathcal{T}$ and $ t^{'} \sim \mathcal{T}$;
		\State Generate two views $\mathcal{G}^{'}=t(\mathcal{G})$ and  $\mathcal{G}^{'}_{T}=t^{'}(\mathcal{G}_{T}$);
		\State Obtain node embeddings $ \mathbf{Z}_{1} $ of $ \mathcal{G}^{'} $
		\State Obtain node embeddings $ \mathbf{Z}_{2} $ of $ \mathcal{G}^{'}_{T} $
		
		\If {$E<T^{'}$} \State Computer the contrastive objective $\mathcal{L}_{info}  $ with Eq. (\ref{l1})
		\State  Update trained parameters in encoder $ f $ with gradient ascent to minimize $\mathcal{L}_{info}  $
		\Else \State Computer the refined contrastive objective $\mathcal{L}  $ with Eq. (\ref{l2}), Eq.(\ref{b})
		\State  Update trained parameters in encoder $ f $ with gradient ascent to minimize $\mathcal{L}  $ 
		\EndIf
		\EndFor
	\end{algorithmic}
\end{algorithm}

\subsection{The Contrastive Learning Framework}
%The framework of the proposed method \textbf{GraphTP} is similar to the general graph contrastive learning method. 
%That is, after augmenting the data, for each node, the distance between the positive sample in the representation space is shortened, and the negative samples are vice versa. 
%The special areas in our method are that the obtained contrastive view has more internal information in the graph, and the selected negative instances are more accurate. 
Based on the designed data augmentations in Section 4.2 and the new negative instance selection strategy in Section 4.3,
the procedure of our model GraphTP is detailed in Algorithm 1.

In summary, given a graph $ \mathcal{G}(\mathbf{X},\mathbf{A}) $, we can obtain its topology augmented graph $ \mathcal{G}_{T}(\mathbf{X},\mathbf{A}_{T}) $ by Sec.\ref{subsec:topology}. Furthermore, we define the set of typical  perturbations as $ \mathcal{T} $. 
Within every epoch, we sample two perturbations $ t \sim \mathcal{T}$ and  $ t^{'} \sim \mathcal{T} $ 
to generate two views: $ \mathcal{G}^{'}=t(\mathcal{G}) $  and $ \mathcal{G}^{'}_{T}=t^{'}(\mathcal{G} _{T})$. 
Then the two views are input into the encoder $ f $ to get the node embeddings: $ \mathbf{Z}_{1} $ and  $ \mathbf{Z}_{2} $. At this time, for each node $ v_{i} $ in the view $ \mathcal{G}^{'} $, its corresponding node $ v_{i}^{'}  $  in another view $ \mathcal{G}^{'}_{T} $ is regarded as the positive. 
While its set of negative instances  $ \mathcal{B}_{f}(v_{i}) $ can be obtained by Sec. \ref{subsec:GCL}. As a result, the contrastive loss of each pair$ ( v_{i}, v_{i}^{'})  $ is formulated as follow:
\begin{equation}
	{l}_{f}(\mathbf{z}_{i}, \mathbf{z}_{i}^{'})=log\frac{e^{\theta(\mathbf{z}_{i},\mathbf{z}_{i}^{'})/\tau}}{\sum_{\mathbf{z}_{j} \in {\mathbf\{{z}_{i}^{'}} \cup \mathcal{B}_{f}(v_{i})\}}e^{\theta(\mathbf{z}_{i},\mathbf{z}_{j})/\tau}}
\end{equation}
where $ \tau $ is the temperature parameter, $ \mathbf{z}_{i} \in \mathbf{Z}_{1}$ and $ \mathbf{z}_{i}^{'} \in \mathbf{Z}_{2} $ are the embeddings of nodes $ v_{i}$ and $ v_{i}^{'}$. $ \theta(\mathbf{z}_{i},\mathbf{z}_{i}^{'})=\mathbf{z}_{i}^\top\mathbf{z}_{i}^{'} $ is the similarity of the pair. For the another view, its contrastive loss can similarly be defined. At last, the overall objective of all $ N $ nodes to be minimized is defined as:
\begin{equation}
	\mathcal{L}=-\frac{1}{2N}\sum^{N}_{i=1}({l}_{f}(\mathbf{z}_{i}, \mathbf{z}_{i}^{'})+l_{f}(\mathbf{z}_{i}^{'},\mathbf{z}_{i}))
	\label{l2}
\end{equation}
In this case, minimizing the pairwise objective can also be seen as maximizing the classical triplet loss:

\begin{equation}
	\begin{split}
		- 	{l}_{f}(\mathbf{z}_{i}, \mathbf{z}_{i}^{'}) \propto 
		 2\tau N_{f} + \sum_{\mathbf{z}_{j} \in  \mathcal{B}_{f}(v_{i})} (  \| {\mathbf{z}_{i}} - {\mathbf{z}_{i}^{'}} \|^2 - \| {\mathbf{z}_{i}} - {\mathbf{z}_{j}} \|^2).
	\end{split}
\end{equation}
We first rearrange the pairwise objective as:
\begin{equation}
	\begin{aligned}
		-	{l}_{f}(\mathbf{z}_{i}, \mathbf{z}_{i}^{'}) & = - log\frac{e^{(\mathbf{z}_{i}^\top\mathbf{z}_{i}^{'})/\tau}}{\sum_{\mathbf{z}_{j} \in {\mathbf\{{z}_{i}^{'}} \cup \mathcal{B}_{f}(v_{i})\}}e^{(\mathbf{z}_{i}^\top\mathbf{z}_{j})/\tau}}\\
		& = log ( 1 + {\sum_{\mathbf{z}_{j} \in  \mathcal{B}_{f}(v_{i})}e^{(\mathbf{z}_{i}^\top\mathbf{z}_{j}-\mathbf{z}_{i}^\top\mathbf{z}_{i}^{'}) /\tau}} ).
	\end{aligned}
\end{equation}
By Taylor expansion of first order, the main derivation process is as follows:
\begin{equation}
	\begin{aligned}
		& - 	{l}_{f}(\mathbf{z}_{i}, \mathbf{z}_{i}^{'}) \\
		& \approx{\sum_{\mathbf{z}_{j} \in  \mathcal{B}_{f}(v_{i})}exp({\frac{\mathbf{z}_{i}^\top\mathbf{z}_{j}-\mathbf{z}_{i}^\top\mathbf{z}_{i}^{'}}{\tau} )}}  \\
		& \approx 1+ \frac 1 \tau\sum_{\mathbf{z}_{j} \in  \mathcal{B}_{f}(v_{i})}({{\mathbf{z}_{i}^\top\mathbf{z}_{j}-\mathbf{z}_{i}^\top\mathbf{z}_{i}^{'}} )} \\
		& = 1 - \frac 1 {2 \tau}\sum_{\mathbf{z}_{j} \in  \mathcal{B}_{f}(v_{i})} ( \| {\mathbf{z}_{i}} - {\mathbf{z}_{j}} \|^2 - \| {\mathbf{z}_{i}} - {\mathbf{z}_{i}^{'}} \|^2) , \\
		& \propto 2\tau N_{f} + \sum_{\mathbf{z}_{j} \in  \mathcal{B}_{f}(v_{i})} (  \| {\mathbf{z}_{i}} - {\mathbf{z}_{i}^{'}} \|^2 - \| {\mathbf{z}_{i}} - {\mathbf{z}_{j}} \|^2) , \\
	\end{aligned}
\end{equation}
which concludes the proof, where $ N_{f} $ is the number of nodes in $ \mathcal{B}_{f}(v_{i}) $.

%s(\mathbf{z}_{i},c)
%_{i=1}^{K}
%\subsubsection{}
%\subsubsection{Theoretical Analysis}

\section{EXPERIMENT}
In this section, we conduct extensive experiments to evaluate our model on five datasets. In particular, we focus on node-level representation learning, where downstream tasks include node classification, clustering, and similarity search.
\subsection{Datasets}
To verify the effectiveness of the model, we take five widely used benchmark datasets of different sizes collected from real networks, including citation networks (CiteSeer, Coauthor-CS) and social networks (WikiCS, Amazon-Computer, Amazon-Photo). The detailed descriptions are given in Table \ref{dataset}.
%Among them,

$ \bullet $ \textbf{CiteSeer} and \textbf{Coauthor-CS } are two citation networks, the nodes of citeseer are represented as publications, and the edges represent citations.

$ \bullet $ \textbf{WikiCS}  is a computer-science related network built on Wikipedia. Nodes are articles and labeled with ten classes. Edges are hyperlinks between articles. Nodes  The features of the nodes are the average of the pre-trained embeddings of the words in each article.

$ \bullet $ \textbf{Amazon-Computers }and \textbf{Amazon-Photo } are two co-purchasing relationship networks, where nodes are goods, and when two goods are often purchased together, an edge is constructed between them. 

\begin{table}[t]
	\centering
	\tiny
	\setlength{\tabcolsep}{1.0mm}
	\caption{The statistics of five benchmark datasets.}
	\begin{tabular}{crrrr}
		\toprule
		Dataset & \multicolumn{1}{c}{\# Nodes} & \multicolumn{1}{c}{\#Edges} & \multicolumn{1}{c}{\#Features} & \multicolumn{1}{c}{\#Labels} \\
		\midrule
		\midrule
		CiteSeer & 3,327 & 4,552 & 3,703 & 6 \\
		WikiCS & 11,701 & 216,123 & 300 & 10 \\
		Coauthor-CS & 18,333 & 81,894 & 6,805 & 15 \\
		Amazon-Photo & 7,650 & 119,081 & 745 & 10 \\
		Amazon-Computers & 13,752 & 245,861 & 767& 10 \\
		\bottomrule
	\end{tabular}%
	\label{dataset}%
\end{table}%
\subsection{Baselines}
To verify the effectiveness of the proposed method GraphTP, for the node classification task, we select representative baselines similar to this paper, which include the following: 

$ \bullet $ Deepwalk \cite{perozzi2014deepwalk}, an unsupervised random-walk model. 

$ \bullet $ DGI \cite{velickovic2019deep} (Deep graph Infomax) applies the Infomax criterion and the augmentation of shuffling node features to develop a GCL proposal.

$ \bullet $ GMI  \cite{peng2020graph} (Graph Mutual Information) further improves DGI by using discriminators to measure the mutual information between the input and the representation of node and edge, respectively.

$ \bullet $ GBT \cite{bielak2022graph} (Graph Barlow Twins) proposes a cross-correlation-based loss objective.

$ \bullet $ MVGRL \cite{hassani2020contrastive} (Multi-View Graph Representation Learning) introduces the graph diffusion technology for multi-view graph contrastive learning. 

$ \bullet $ GCA \cite{zhu2021graph} proposes the method of adaptive data augmentation with prior knowledge.

$ \bullet $ GRACE \cite{zhu2020deep} designs various graph data augmentations, such as removing edges and masking node features.

$ \bullet $ COSTA \cite{zhang2022costa} proposes the feature augmentation from the perspective of covariance. 

%$ \bullet $ ENGAGE \cite{shi2023engage} 

$ \bullet $  HomoGCL \cite{li2023homogcl}  is a recent work that mines neighboring nodes with special significance for nodes to expand the positive set.

Furthermore, we also report the performance of Graph Convolutional Networks (\textbf{GCNs}) \cite{welling2016semi} under fully supervised conditions, trained in an end-to-end manner. 
In addition, in order to verify the generalization of the model, the node clustering and the similarity search are introduced. The BGRL \cite{thakoor2021bootstrapped} and AFGRL \cite{lee2022augmentation} are further used to compare.
The former method designs a framework for contrastive learning without negative samples. The latter goes a step further by introducing data-free augmentation and more positive samples for representation learning. For all baselines, we report the performance according to their official implementation.
\subsection{Implementation Details}
In our experiment, we adopt a two-layer GCNs \cite{welling2016semi} as the backbone network. 
All experiments are implemented by using PyTorch and optimized with the Adam optimizer. For hyper-parameter settings, 
the embedding dimension is set to 256 for all datasets except Amazon-Computers where $ d^{'}=128 $. The learning rate is set to 0.01 for Amazon-Photo, Amazon-Computers, and WikiCS, 0.001 for CiteSeer, and 0.0005 for Coauthor-CS. The setting of top-$k$ in the scheme of feature space and matrix transformation is 1 for CiteSeer and WikiCS, while $ k=10 $ for two datasets of Amazon and $ k=12 $ for Coauthor-CS. 
Following the order of datasets in Table \ref{dataset}, the parameter $ \alpha$ is set to 180, 100, 160, 80, and 80. Meanwhile, in the operation of building the prototype, $ K $ is uniformly set to $ 100 $ for all datasets. 

For each experiment, the model is firstly trained in an unsupervised manner by adopting the designed method. Then for the task of node classification, the resulting embeddings are fed to a $ l_{2} $-regularized logistic regression classifier to evaluate. The training set, validation set, and test set in the experiment are all divided based on the settings of previous work \cite{zhu2021graph,zhang2022costa}.
For node clustering and similarity search, the resulting embeddings are directly used to evaluate. 

\subsection{Experimental Results}
% Table generated by Excel2LaTeX from sheet 'Sheet1'
\begin{table*}[htbp]
	\centering
	\caption{Node classification result in terms of average accuracy($\uparrow $) in percentage with standard deviation. The second column represents the available data for each model during training, where $ \mathbf{X} $,$ \mathbf{A} $,$ \mathbf{Y} $  correspond to node features, the adjacency matrix and labels separately. }
			\tiny
	\begin{tabular}{cccccccc}
		\toprule
		\textbf{Method} & \textbf{Training Data} & \textbf{Amazon-Photo} & \textbf{Amazon-Computers} & \textbf{Coauthor-CS}& \textbf{Wiki-CS}  & \textbf{CiteSeer}  \\
		\midrule
		\midrule
		GCN   & $ \mathbf{X} $,$ \mathbf{A} $,$ \mathbf{Y} $ & 92.42 ± 0.22 & 86.51 ± 0.54 & 93.03 ± 0.31 & 77.19 ± 0.12  & 70.4 ± 0.4 \\
		DeepWalk & $ \mathbf{A} $     & 89.44 ± 0.11 & 85.68 ± 0.06 & 84.61 ± 0.22   & 74.35 ± 0.06 & 50.5 ± 0.2 \\
		Raw features & $ \mathbf{X} $    & 78.53 ± 0.00 & 73.81 ± 0.00 & 90.37 ± 0.00   & 71.98 ± 0.00 & 64.6 ± 0.0 \\
		\midrule
		GBT & $ \mathbf{X} $,$ \mathbf{A} $   & 92.46 ± 0.35 & 87.93 ± 0.36 & 92.91 ± 0.25 & 76.83 ± 0.73 & 69.4 ± 0.5  \\
		DGI   & $ \mathbf{X} $,$ \mathbf{A} $  & 91.61 ± 0.22 & 83.95 ± 0.47 & 92.15 ± 0.63 & 75.35 ± 0.14 & 68.8 ± 0.7  \\
		GMI   & $ \mathbf{X} $,$ \mathbf{A} $  & 90.68 ± 0.17 & 82.21 ± 0.31 & 91.08 ± 0.56  & 74.85 ± 0.08  & 72.4 ± 0.1 \\
		GCA   & $ \mathbf{X} $,$ \mathbf{A} $  & 92.49 ± 0.09 & 87.85 ± 0.31 & 93.10 ± 0.01 & 78.30 ± 0.00 & 71.5 ± 0.3  \\
		MVGRL & $ \mathbf{X} $,$ \mathbf{A} $ & 91.74 ± 0.07 & 87.52 ± 0.11 & 92.11 ± 0.12 & 77.52 ± 0.08 & 72.2 ± 1.3  \\
		GRACE & $ \mathbf{X} $,$ \mathbf{A} $   & 92.53 ± 0.16 & 87.80 ± 0.23 & 92.95 ± 0.03 & 78.31 ± 0.05 & 72.1 ± 0.5  \\
		HomoGCL & $ \mathbf{X} $,$ \mathbf{A} $   & 92.92 ± 0.18 & 88.46 ± 0.20 & 92.16 ± 0.05 & 78.26 ± 0.21 & 72.3 ± 0.7  \\
		COSTA\_SV & $ \mathbf{X} $,$ \mathbf{A} $  & 92.30 ± 0.25 & 88.26 ± 0.03 & 92.95 ± 0.12  & 79.03 ± 0.05 & 72.8 ± 0.3  \\
		COSTA\_MV & $ \mathbf{X} $,$ \mathbf{A} $   &92.56 ± 0.45 & 88.32 ± 0.03 & 92.94 ± 0.10  &\textbf{79.12 ± 0.02}  & 72.9 ± 0.3 \\
		\midrule
		GraphTP-T & $ \mathbf{X} $,$ \mathbf{A} $  & \textbf{93.72 ± 0.14} & \underline{89.54 ± 0.18} & \underline{93.30 ± 0.01 }& \underline{79.07 ± 0.03} & \textbf{74.1 ± 0.2 } \\
		GraphTP-F & $ \mathbf{X} $,$ \mathbf{A} $   & \underline{93.69 ± 0.13 }& \textbf{89.93 ± 0.14} & \textbf{93.81 ± 0.02} & 79.01 ± 0.04 & \underline{73.8 ± 0.3}  \\
		\bottomrule
	\end{tabular}
	\vspace{0.6em}
	\label{node}%
\end{table*}%
\begin{table}[t!]
	\centering
	\tiny
	\setlength{\tabcolsep}{0.8mm}
	\caption{Performance on node clustering in terms of NMI$\uparrow $ and Hom$\uparrow $(homogeneity).}
	\begin{tabular}{c|c|cccc|c}
		\multicolumn{2}{c|}{\textbf{Methods}} & \textbf{GRACE} & \textbf{GCA} & \textbf{BGRL} & \textbf{AFGRL} & \textbf{GraphTP} \\
		\midrule
		\midrule
		\multirow{2}[2]{*}{\textbf{WikiCS}} & \textbf{NMI} & \underline{0.4282} & 0.3373 & 0.3969 & 0.4132 & \textbf{0.4634} \\
		& \textbf{Hom} &\underline{ 0.4423} & 0.3525 & 0.4156 & 0.4307 &\textbf{ 0.4821} \\
		\midrule
		\multirow{2}[2]{*}{\textbf{Computers}} & \textbf{NMI} & 0.4793 & 0.5278 & 0.5364 & \underline{0.5520} & \textbf{0.5765} \\
		& \textbf{Hom} & 0.5222 & 0.5816 & 0.5869 &\underline{ 0.6040} & \textbf{0.6353} \\
		\midrule
		\multirow{2}[2]{*}{\textbf{Photo}} & \textbf{NMI} & 0.6513 & 0.6443 &\underline{ 0.6814} & 0.6563 & \textbf{0.7183} \\
		& \textbf{Hom} & 0.6657 & 0.6575 & \underline{0.7004} & 0.6743 & \textbf{0.7362} \\
		\midrule
		\multirow{2}[2]{*}{\textbf{Coauthor-CS}} & \textbf{NMI} & 0.7562 & 0.762 & 0.7732 & \underline{0.7859} & \textbf{0.7901} \\
		& \textbf{Hom} & 0.7909 & 0.7965 & 0.8041 & \underline{0.8161} & \textbf{0.8213} \\
		\bottomrule
	\end{tabular}%
	\label{cluster}%
\end{table}%

%\subsubsection{Performance Comparison}
The experimental result of node classification is shown in Table \ref{node}. 
Overall, GraphTP achieves competitive (first or second place) performance compared to state-of-the-art algorithms on benchmark datasets, which verifies the effectiveness of our method.
Specifically, GraphTP improves performance by 1.61\%, 1.16\%, and 1.2\% over the best baseline on Amazon-Photo, Amazon-Computer, and CiteSeer, respectively. On the Coauthor-CS, we find that although existing baselines have achieved sufficiently high performance, our approach still pushes the frontier in accuracy by almost 1\%.  And there are not many behind on the WikiCS with SOTA, only 0.05\%.

Meanwhile, we observe other observations as follows. 
The shallow methods including Raw feature and DeepWalk performed worse, which cannot both utilize the information of raw features and adjacency matrix. The early contrastive learning methods DGI and GMI also do not show competitive performance. They focus on modeling the whole graph or subgraph structure after simple or no data augmentations. Compared with the methods that focus on graph data augmentations (GRACE, GCA, MVGRL),
%the two augmentation schemes we propose can provide a more sufficient and informative contrastive view, thereby improving the performance.
the excellent performance of GraphTP verifies that our two proposed topology augmented schemes for graph data can help improve the quality of representation learning.
Although MVGRL adopts the method of injecting external information into the augmented view, it is still fair to augment the important edges more strongly on the input graph or take use of the topology information inside the feature space. 
Compared with the recent methods: HomoGCL with the use of homophily assumption and COSTA with feature-level augmentation to alleviate the problem of biased augmentation, our design of combining the topology reorganization and prototype-based selective sampler is more effective. 
%Moreover, GraphTP is also competitive with supervised GCN methods without label guidance.
\begin{table}[t!]
	\centering
	\tiny
	\setlength{\tabcolsep}{0.8mm}
	\caption{Performance on similarity search. (Sim@$ n $$\uparrow $: the average ratio between $ n $ nearest neighbors that share the same label as the query node.)}
	\begin{tabular}{c|c|cccc|c}
		\multicolumn{2}{c|}{\textbf{Methods}} & \textbf{GRACE} & \textbf{GCA} & \textbf{BGRL} & \textbf{AFGRL} & \textbf{GraphTP} \\
		\midrule
		\midrule
		\multirow{2}[2]{*}{\textbf{WikiCS}} & \textbf{sim@5} & 0.7754 & 0.7786 & 0.7739 & \underline{0.7811} & \textbf{0.7841} \\
		& \textbf{sim@10} & 0.7645 & 0.7673 & 0.7617 & \underline{0.7660} & \textbf{0.7753} \\
		\midrule
		\multirow{2}[2]{*}{\textbf{Computers}} & \textbf{sim@5} & 0.8738 & 0.8826 & 0.8947 & \textbf{0.8966} & 0.8918 \\
		& \textbf{sim@10} & 0.8643 & 0.8742 & 0.8855 & \textbf{0.8890} & 0.8851 \\
		\midrule
		\multirow{2}[2]{*}{\textbf{Photo}} & \textbf{sim@5} & 0.9155 & 0.9112 & \underline{0.9245} & 0.9236 & \textbf{0.9272} \\
		& \textbf{sim@10} & 0.9106 & 0.9052 & \underline{0.9195} & 0.9173 & \textbf{0.9225 }\\
		\midrule
		\multirow{2}[2]{*}{\textbf{Coauthor-CS}} & \textbf{sim@5} & 0.9104 & 0.9126 & 0.9112 & \underline{0.9180} & \textbf{0.9205} \\
		& \textbf{sim@10} & 0.9059 & 0.9100  & 0.9086 & \underline{0.9142} & \textbf{0.9182 }\\
		\bottomrule
	\end{tabular}%
	\label{tab:sim}%
\end{table}%
We also evaluate the performance on the tasks of node clustering (Table \ref{cluster}) and similarity search (Table \ref{tab:sim}), where adopts the scheme 1 to genetare the contrastive view. Table \ref{cluster} shows that GraphTP generally outperforms other methods in node clustering. Among them, the highest improvement in clustering indicators are 8.2\% and 8.9\%.
We think that this is mainly because GraphTP is different from other contrastive methods in the selection of negative samples. False negative samples are screened out based on semantic information in GraphTP so that the clusters formed by clustering are tighter and the distance between clusters is larger. For a more intuitive display, we make a visual description later in Sec. \ref{vis}.
Meanwhile, our method performs well in terms of node similarity search. In the baselines, AFGRL designs the strategy of treating the nearer semantic neighbor of the anchor as positive samples, which is very beneficial to this task, but our method is still superior to it on the three datasets. 

In short, compared with the existing advanced methods on three tasks, the performance verifies the effectiveness of our proposed framework. %and it is a more thoughtful contrastive learning approach.

\begin{figure*}
	\begin{center}
		\includegraphics[scale=0.44]{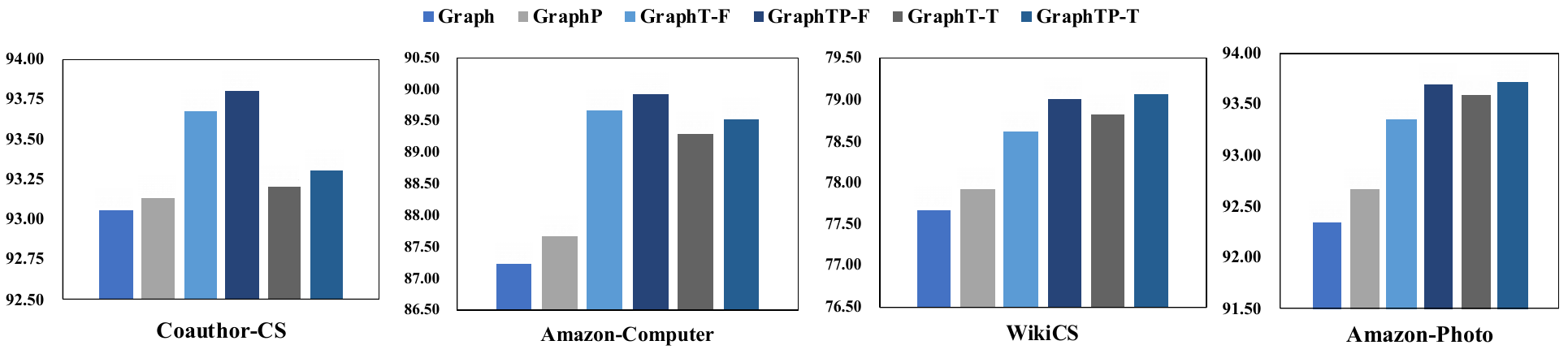}
		\caption{Ablation Study.}
		\label{fablation}
	\end{center}	
\end{figure*}	
% Table generated by Excel2LaTeX from sheet 'Sheet1'	

\begin{table}[t!]
	\centering
	\setlength{\tabcolsep}{0.6mm}
	\tiny
	\caption{Explanation of variants, where T-1 uses scheme 1 for topology augmentation in Sec. \ref{subsec:topology}, T-2 uses scheme 2; Selection corresponds to filtering negative samples.}
	\begin{tabular}{c|cccccc}
		\toprule
		\multicolumn{1}{c}{Variants} & Graph & GraphP & GraphT-F & GraphTP-F & GraphT-T & GraphTP-T \\
		\midrule
		\midrule
		T-1 &  \ding{55}     &  \ding{55}     & \checkmark      &\checkmark      & -     & - \\
		T-2 &  \ding{55}    &  \ding{55}     & -     & -     & \checkmark     & \checkmark  \\
		Selection & \ding{55}     & \checkmark     &  \ding{55}     & \checkmark      &  \ding{55}    & \checkmark  \\
		\bottomrule
	\end{tabular}%
	\label{tab:ab}%
\end{table}%

\begin{table}[t!]
	\centering
	\caption{ Performance with different $ K $.}
	\tiny
	\begin{tabular}{c|cccc}
		\toprule
		\multicolumn{1}{c}{} & $ K $=10  & $ K $=50  & $ K $=100 & $ K $=200 \\
		\midrule
		\midrule
		CiteSeer & 73.63 & 73.91 & 73.82 & 73.83 \\
		WikiCS & 78.96 & 78.95 & 79.01 & 78.91 \\
		Coauthor-CS & 93.77 & 93.78 & 93.81 & 93.72 \\
		Amazon-Photo & 93.33 & 93.54 & 93.69 & 93.46 \\
		Amazon-Computer & 89.87 & 89.91 & 89.93 & 89.92 \\
		\bottomrule
	\end{tabular}%
	\label{tab:kmeas}%
\end{table}%

\subsection{Ablation Study}
To further study the effectiveness of each part in the designed model, we conduct ablation experiments on the datasets. In a word, the innovations of model are mainly in two components, namely data augmentation by \textbf{T}opology reorganization and \textbf{P}rototype-based negative sample selection. And there are two schemes in data augmentation. 
Therefore, we first remove the two components in the model and name it \textbf{Graph}, then two topology reorganization schemes are added separately and defined as two variants, namely \textbf{GraphT-F }and \textbf{GraphT-T}. To verify the effectiveness of the proposed negative sample selection strategy, we further added the strategy to the first three variants, named\textbf{ GraphP}, \textbf{GraphTP-F}, and \textbf{GraphTP-T}, respectively. The corresponding instructions are given in Table \ref{tab:ab} . 

\begin{figure}[t!]
	\begin{center}
		\includegraphics[scale=0.35]{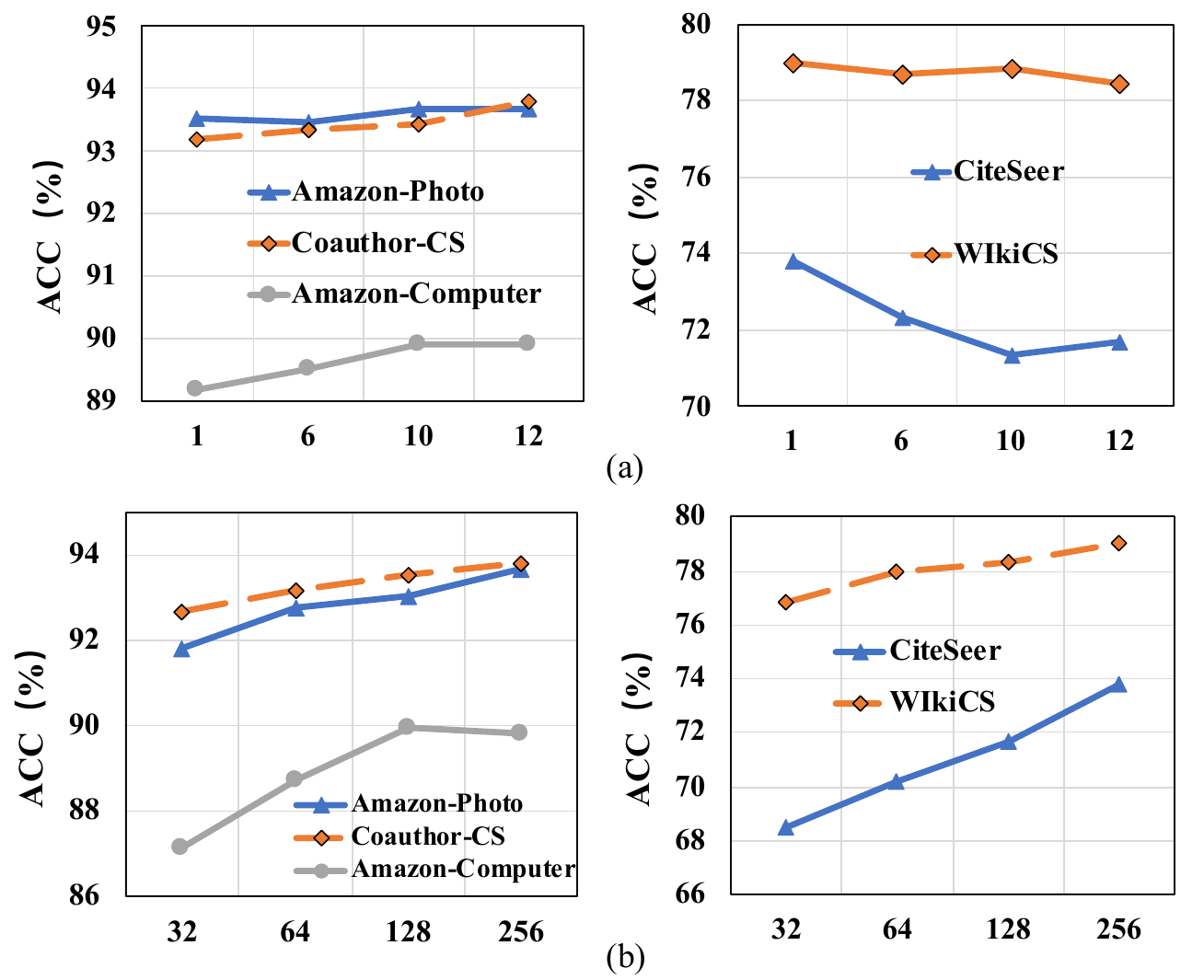}
		\caption{The experiment result of the parameter study.}
		\label{em}
	\end{center}	
\end{figure}

The results are shown in Figure. \ref{fablation}, where we can see that GraphTP-T and GraphTP-F consistently outperform the other variants on all datasets, which indicates that both components contribute to the investigated tasks. 
Among them, we firstly find that the performance of two topology reorganized views Graph\textbf{T-T} and Graph\textbf{T-F }are much better than the basic variant \textbf{Graph}. It validates the need for targeted augmentation of graph structures.
Meanwhile, we find that the variants after adding the negative sample selection strategy: Graph\textbf{P}, GraphT\textbf{P}-F and GraphT\textbf{P}-T have improved performance compared to the previous ones, which verifies the necessity of removing false negative samples. 
In addition, we can observe that the gain brought by topology augmentation is larger than the designed negative sample selection strategy, which reflects the importance of view generation as the first step in the contrastive learning framework. 
GraphT-\textbf{F} and its variants outperform GraphT-\textbf{T} on Coauthor-CS and Amazon-Computer, while the opposite is true on the other datasets. %The reason may be that the effect of different modules on datasets varies due to the data characteristics, where the information insides the feature space is more useful for them.
% Table generated by Excel2LaTeX from sheet 'Sheet1'

\begin{figure*}
	\begin{center}
		\includegraphics[scale=0.5]{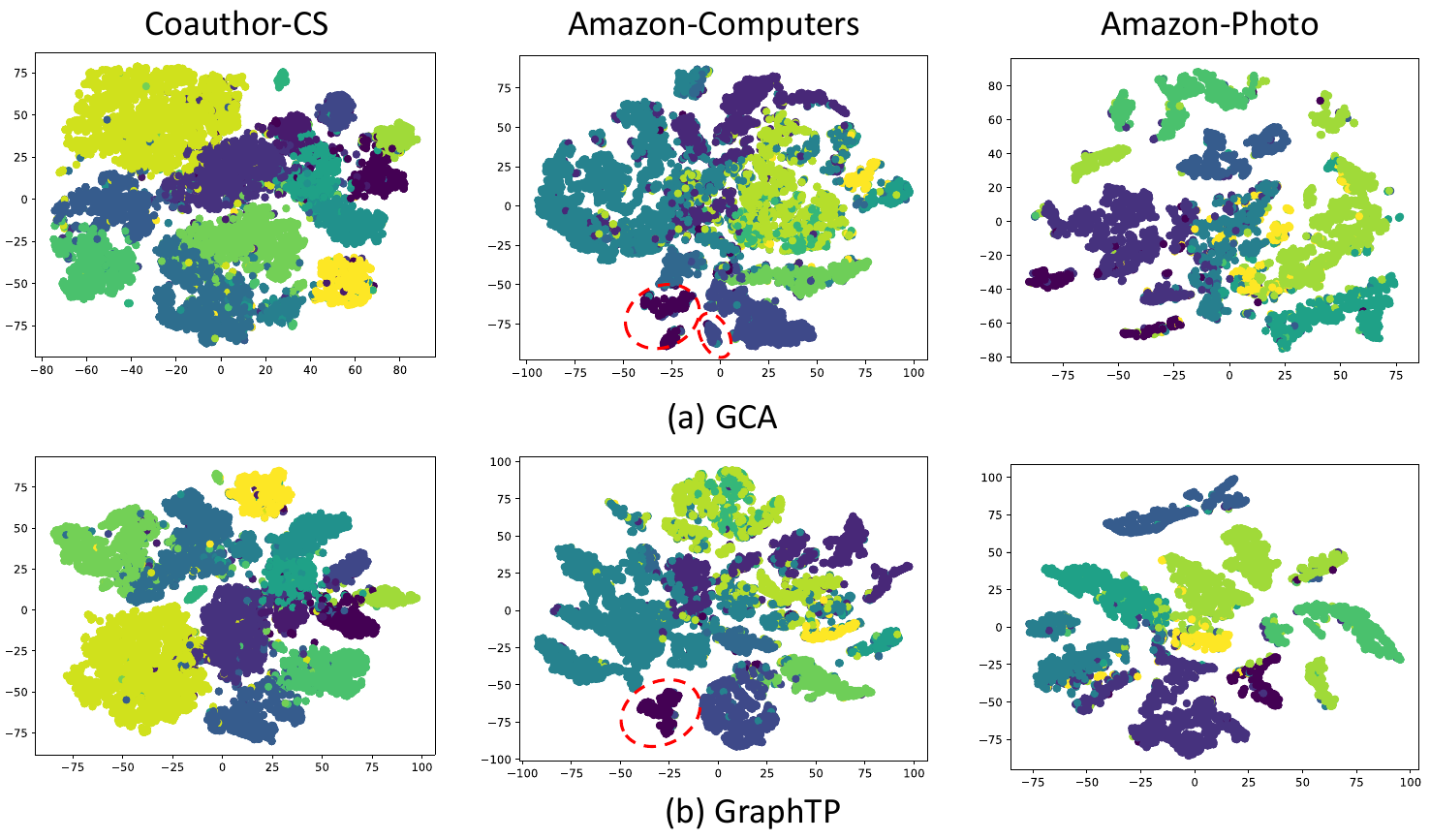}
		\caption{Visualization of the learned node embeddings of GraphTP and GCA on three datasets.}
		\label{emm}
	\end{center}	
\end{figure*}		

\subsection{Parameter Study}
In this subsection, we investigate the effect of three hyperparameters in our proposed model on five benchmark datasets, which are the number of clusters $ K $ when selecting negative samples based on prototypes, the hidden size $ d^{'} $ in the encoder, and the number of neighbors $ k $ selected based on similarity after topology reorganization with scheme 1.

1) In this experiment, we change the value of $ k  $ (1,6,10,12) to understand how this parameter affects the performance of GraphTP. As shown in Figure 5(a), $ k $ has different effects on the performance of each dataset. A too-small or large value of $ k $ will affect their performance. 
Among them, the Coauthor-CS with the largest number of nodes needs a larger $ k $ value, that is, $ k $=12. Citeseer is affected to a greater degree, and it will be optimal when $ k $=1.

2) We study the sensitivity of the hidden size $ d^{'}$ for the proposed framework GraphTP. Here, we choose 32, 64, 128, and 256 as the hidden size. 
%We aims to explore how increasing $ d^{'}$ affects the accuracy of GraphTP on benchmark datasets. 
From the Figure 5(b), we can see that all datasets have a relatively obvious upward trend, which indicates that $ d^{'}$ is positively correlated with the performance of GraphTP within a certain range. The reason behind this may be that increasing $ d^{'}$ will increase the number of trainable parameters. In addition, the accuracy of some datasets is in a state of continuous improvement in the selected parameters. But in order to the comprehensive performance and operating efficiency, while referring to the parameter settings of previous work, we finally chose $ d^{'}$=256 for these data sets, while naturally set it to 128 for the Amazon-Computers.

3) To explore the importance of $ K $ on different datasets, we conduct experiments with four different values of $ K  $ (10, 50, 100, and 200, respectively). Our results are shown in Table \ref{tab:kmeas}, in which the accuracy is relatively stable on all datasets and is less affected by $ K $. Most of them achieve the best results when $ K $=100. The reason why CiteSeer is slightly better than 100 on 50 may be that its number of nodes is less than the other four datasets. In the end, we all uniformly set it to 100 
for convenience.

\subsection{Visualization}
\label{vis}
To demonstrate the learned node embedding and display the benefits of methods more intuitively, we visualize the node embeddings of GCA and GraphTP on the Amazon-Photo, Amazon-CS, and Coauthor-CS datasets. %Specifically, we use the embedding of the last layer in the graph encoder and t-SNE for feature visualization. Each point represents a node when the color represents the node label. 
%Then the node embeddings generated by the two methods are grouped according to their corresponding node labels. 
From the Figure \ref{emm}, it can be observed that the embeddings learned by GraphTP have better clustering results, in which the clusters are more compact. In detail, GraphTP can capture finer-grained category semantic information, and there are more obvious boundaries between different node clusters. 

In particular, it is more intuitive to verify that our method can effectively alleviate the risk of semantic drift from the visualization in the Amazon-Computers. In the lower part of the Figure 6(a), we can find that in GCA, the clusters of the same class are separated, and the separated part moves to other classes. This is because under the trivial negative sample selection, instances of the same category with anchor are considered as negative and mistakenly excluded.
In contrast, our method can greatly alleviate this phenomenon by removing these false negative instances. Thus, the intra-class distance in GraphTP is tighter. 
%Meanwhile, the quantitative result of the clustering effect is given in Table \ref{cluster}, which also reflects the effectiveness of our proposed method.
\section{CONCLUSION}
In this paper, we develop a novel topology augmentation and a new negative sample selection strategy for node representation learning in graph, which is named GraphTP. In our model, the contributions of GraphTP consists of three parts. First, we focus on the topology relationship of the graph to augment globally, which takes use of the adjacency matrix and the semantic information in feature space.
Second, we further design a strategy for the negative sample selection, which could remove the false negative samples to mitigate the risk of semantic shift.
At last, we conduct comprehensive experiments on various real-world datasets. Experimental results show that GraphTP outperforms most existing state-of-the-art methods, even surpassing supervised methods.

\section*{ACKNOWLEDGMENTS}
This work is supported by the NSFC under granted No. 62076124 and the A3 Foresight Program of NSFC under granted No. 62061146002.
%% The Appendices part is started with the command \appendix;
%% appendix sections are then done as normal sections
%% \appendix

%% \label{}

%% If you have bibdatabase file and want bibtex to generate the
%% bibitems, please use
%%
%%  \bibliographystyle{elsarticle-num} 
%%  \bibliography{<your bibdatabase>}

%% else use the following coding to input the bibitems directly in the
%% TeX file.

\bibliographystyle{elsarticle-num} 
\bibliography{gcl}

%\subsection*{  } % 让reference和biography之间保留一定的间距
%\setlength\intextsep{0pt} % 对齐photo和introduction
%\begin{wrapfigure}{l}{25mm}
%	\centering
%	\includegraphics[width=1in,height=1.25in,clip,keepaspectratio]{1.png}
%\end{wrapfigure}
%\noindent \textbf{Jiaqiang Zhang} received the B.s. degree in Information and Computing Science in 2020. He is a second year Ph.D.student of Department of Computer Science and Technology in Nanjing University of Aeronautics and Astronautics and a member of ParNec Group. His research interest includes self-supervised learning and graph representation learning..\par

%\hspace*{\fill} % author与author之间空一行

%% Author 2
%\subsection*{  } % 让reference和biography之间保留一定的间距
%\setlength\intextsep{0pt} % 对齐photo和introduction
%\begin{wrapfigure}{l}{25mm}
%	\centering
%	\includegraphics[width=1in,height=1.25in,clip,keepaspectratio]{2.png}
%\end{wrapfigure}
%\noindent \textbf{Songcan Chen} received his BS degree in mathematics from Hangzhou University (now merged into Zhejiang University) in 1983. In 1985, he completed his MS degree in computer applications at Shanghai Jiaotong University and then worked at NUAA in January 1986. There he re- ceived a PhD degree in communication and information systems in 1997. Since 1998, as a full-time professor, he has been with the College of Computer Science \& Technology at NUAA. His research interests include pattern recognition, machine learning and neural computing. He is also an IAPR Fellow.\par

\end{document}